\documentclass[sigconf,anonymous=false]{acmart}
\renewcommand\footnotetextcopyrightpermission[1]{} 
\acmDOI{}
\makeatletter
\renewcommand\@formatdoi[1]{\ignorespaces}
\makeatother
\usepackage{graphicx}
\usepackage{stfloats}
\usepackage{subcaption}
\usepackage{amsmath}
\usepackage{bm}
\usepackage{mathtools}
\usepackage{color}

\usepackage{color}

\newcommand{\dl}{{d_{\mathrm{latent}}}}
\newcommand{\dd}{{d_{\mathrm{data}}}}

\usepackage[nonumberlist, acronym, toc]{glossaries}
\newacronym{ml}{ML}{Machine Learning}
\newacronym{es}{ES}{Earth Sciences}
\newacronym{rbm}{RBM}{Restricted Boltzmann Machine}
\newacronym{qml}{QML}{Quantum Machine Learning}
\newacronym{qbm}{QBM}{Quantum Boltzmann Machine}
\newacronym{cd}{CD}{Contrastive Divergence}
\newacronym{pcd}{PCD}{Persistent Contrastive Divergence}
\newacronym{dbn}{DBN}{Deep Belief Net}
\newacronym{qa}{QA}{Quantum Annealing}
\newacronym{qaaan}{QAAAN}{Quantum Assisted Associative Adversarial Network}
\newacronym{mnist}{MNIST}{Modified National Institute of Standards and Technology database}
\newacronym{modis}{MODIS}{Moderate Resolution Imaging Spectroradiometer}
\newacronym{cdt}{CDT}{Centre for Doctoral Training}
\newacronym{gan}{GAN}{Generative Adversarial Network}
\newacronym{sgd}{SGD}{Stochastic Gradient Descent}
\newacronym{qip}{QIP}{Quantum Information Processing}
\newacronym{qc}{QC}{Quantum Computing}
\newacronym{elbo}{ELBO}{Evidence Lower Bound}
\newacronym{vae}{VAE}{Variational Autoencoder}
\newacronym{qvae}{QVAE}{Quantum-assisted Variational Autoencoder}
\newacronym{snail}{SNAIL}{Small Noisy Annealing Implementation Learning}
\newacronym{bm}{BM}{Boltzmann Machine}
\newacronym{kl}{KL}{Kullback-Leibler}
\newacronym{fid}{FID}{Frech\'et Inception Dfistance}
\newacronym{is}{IS}{Inception Score}
\newacronym{lsh}{LSH}{Locality Sensitive Hashing}
\newacronym{anns}{ANNS}{Approximate Nearest Neighbor Search}
\newacronym{hnsw}{HNSW}{Hierarchical Navigable Small World}
\newacronym{nisq}{NISQ}{Noise Intermediate-Scale Quantum}
\newacronym{pmf}{pmf}{probability mass function}
\newacronym{relu}{ReLU}{Rectified Linear Unit}
\newacronym{dvae}{DVAE}{Discrete Variational Autoencoder}
\newacronym{ndvi}{NDVI}{Normalized Vegetation Index}

\AtBeginDocument{%
  \providecommand\BibTeX{{%
    \normalfont B\kern-0.5em{\scshape i\kern-0.25em b}\kern-0.8em\TeX}}}

\begin{document}

\title{High-Dimensional Similarity Search with Quantum-Assisted Variational Autoencoder}

\author{Nicholas Gao}
\orcid{0000-0002-1487-2657}
\affiliation{%
    \department{Quantum Artificial Intelligence Lab.}
    \institution{NASA Ames Research Center}
    \city{Moffett Field}
    \postcode{94035}
    \state{CA}
    \country{USA}
}
\additionalaffiliation{
    \institution{KBR Inc}
    \city{Huston}
    \postcode{77002}
    \state{TX}
    \country{USA}
}
\email{nicholas.gao@tum.de}

\author{Max Wilson}
\affiliation{
    \department{Quantum Artificial Intelligence Lab.}
    \institution{NASA Ames Research Center}
    \city{Moffett Field}
    \postcode{94035}
    \state{CA}
    \country{USA}
}
\authornotemark[1]
\additionalaffiliation{
    \institution{Quantum Engineering CDT, Bristol University}
    \city{Bristol}
    \postcode{BS8 1TH}
    \country{UK}
}
\email{aw16952@bristol.ac.uk}

\author{Thomas Vandal}
\affiliation{%
    \institution{NASA Ames Research Center\\BAER Institute}
    \city{Moffett Field}
    \postcode{94035}
    \state{CA}
    \country{USA}
}
\email{thomas.vandal@nasa.gov}
  
\author{Walter Vinci}
\affiliation{%
    \department{Quantum Artificial Intelligence Lab.}
    \institution{NASA Ames Research Center}
    \city{Moffett Field}
    \postcode{94035}
    \state{CA}
    \country{USA}
}
\authornotemark[1]
\email{walter.vinci@nasa.gov}

\author{Ramakrishna Nemani}
\affiliation{%
    \institution{NASA Ames Research Center}
    \postcode{94035}
    \city{Moffett Field}
    \state{CA}
    \country{USA}
}
\email{rama.nemani@nasa.gov}

\author{Eleanor Rieffel}
\affiliation{%
    \department{Quantum Artificial Intelligence Lab.}
    \institution{NASA Ames Research Center}
    \city{Moffett Field}
    \postcode{94035}
    \state{CA}
    \country{USA}
}
\email{eleanor.rieffel@nasa.gov}

\renewcommand{\shortauthors}{Gao, et al.}

\begin{abstract}
    Recent progress in quantum algorithms and hardware indicates the potential importance of quantum computing in the near future. However, finding suitable application areas remains an active area of research.
    Quantum machine learning \cite{biamonte2017quantum} is touted as a potential approach to demonstrate quantum advantage within both the gate-model \cite{lloyd2014quantum, rebentrost2014quantum} and the adiabatic \cite{mott2017solving, vinci2019path} schemes.
    For instance, the \acrfull{qvae} \cite{khoshaman2018quantum} has been proposed as a quantum enhancement to the discrete \acrshort{vae} ~\cite{rolfe_discrete_2016}. 
    We extend on previous work and study the real-world applicability of a \acrshort{qvae} by presenting a proof-of-concept for similarity search in large-scale high-dimensional datasets.
    While exact and fast similarity search algorithms are available for low dimensional datasets, scaling to high-dimensional data is non-trivial.
    We show how to construct a space-efficient search index based on the latent space representation of a \acrshort{qvae}.
    Our experiments show a correlation between the Hamming distance in the embedded space and the Euclidean distance in the original space on the \acrfull{modis} dataset.
    Further, we find real-world speedups compared to linear search and demonstrate memory-efficient scaling to half a billion data points.
\end{abstract}

\ccsdesc[500]{Theory of computation~Nearest neighbor algorithms}
\ccsdesc[300]{Hardware~Quantum computation}
\ccsdesc[100]{Applied computing~Earth and atmospheric sciences}

\keywords{Similarity Search; Quantum Machine Learning; Variational Autoencoder; Earth Science}


\maketitle
\newcommand*\diff{\mathop{}\!\mathrm{d}}

\section{Introduction} \label{sec:introduction}
\begin{figure}
    \centering
    \begin{subfigure}{\linewidth}
        \centering
        \includegraphics[width=\linewidth]{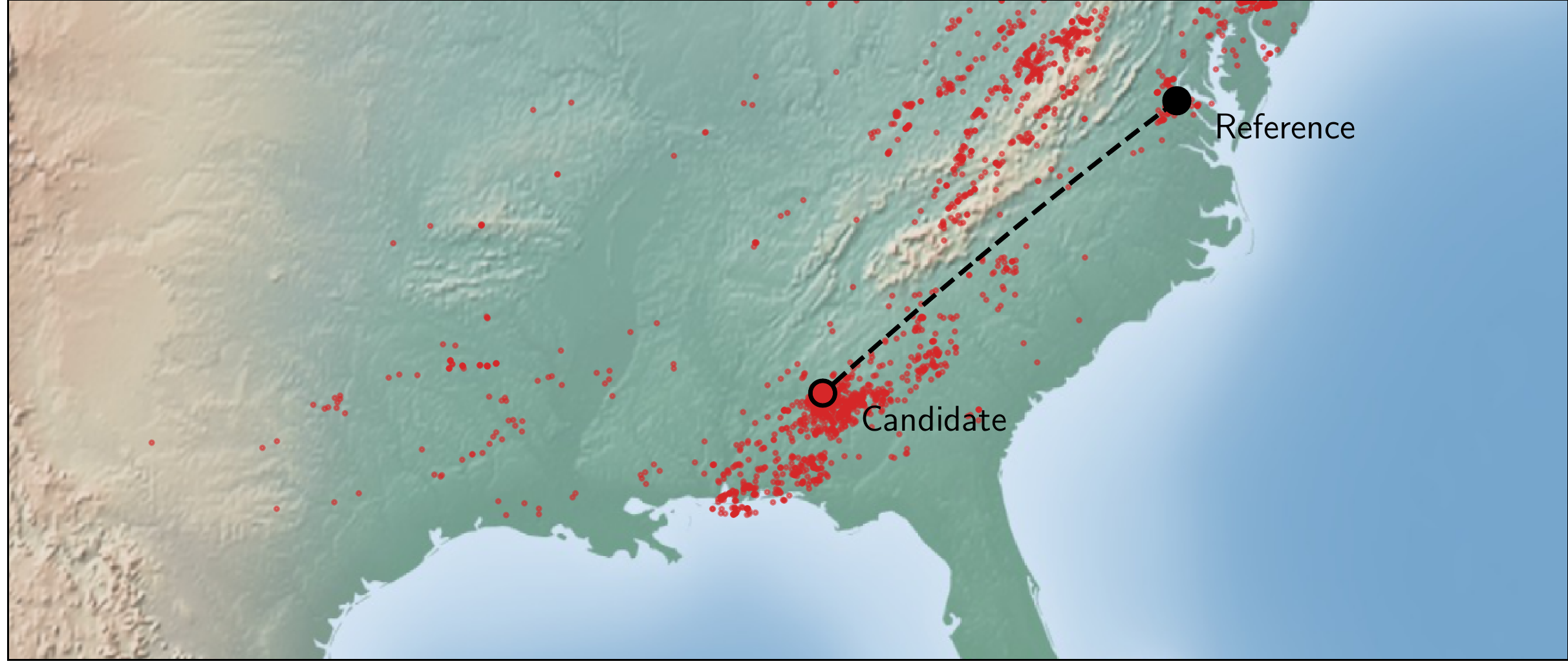}
        \caption{Candidates with Hamming distance $\leq 1$
        }
        \label{fig:ndvi:map}
    \end{subfigure}
    \begin{subfigure}{\linewidth}
        \centering
        \includegraphics[width=\linewidth]{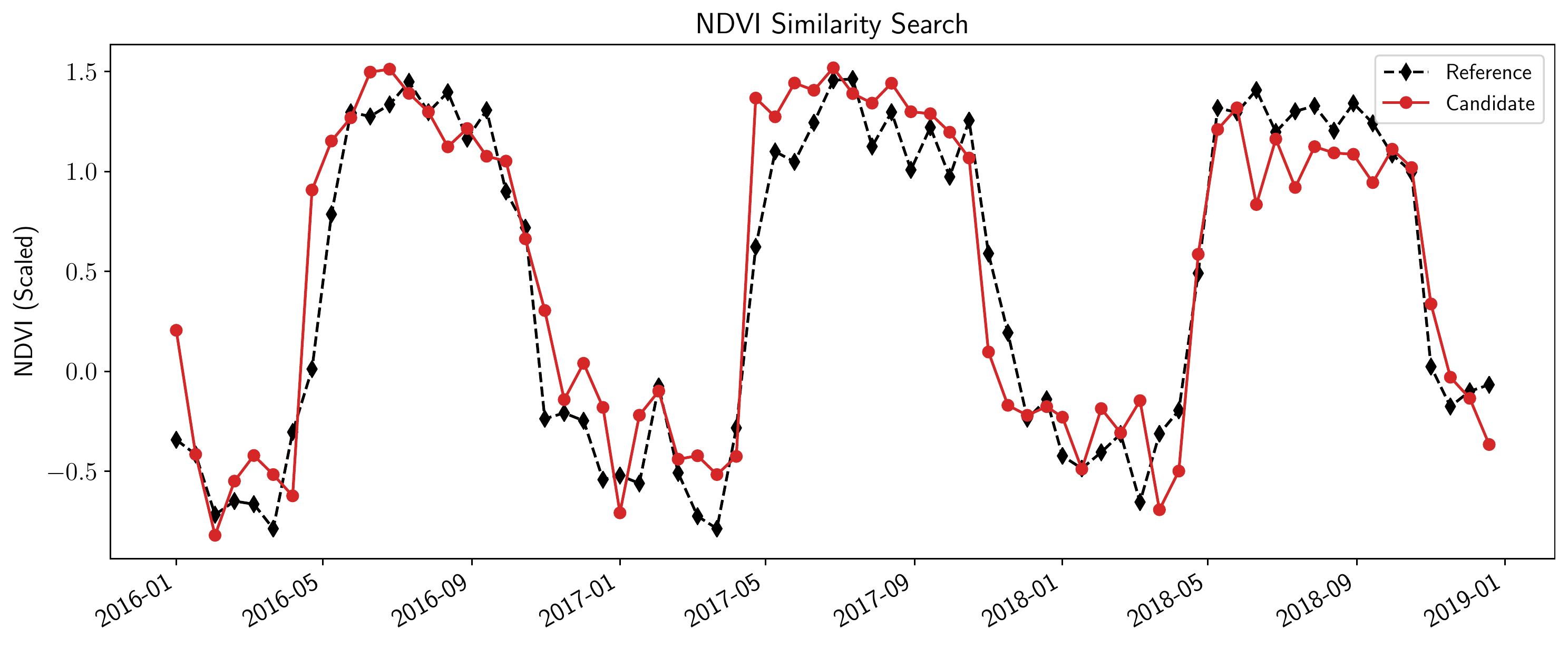}
        \caption{NDVI comparison
        }
        \label{fig:ndvi:comp}
    \end{subfigure}
    \caption{
        Similarity search on the \acrshort{modis} dataset. Figure~\ref{fig:ndvi:map} shows candidates and Figure~\ref{fig:ndvi:comp} shows a comparison between the reference point and a random candidate.
    }
    \label{fig:ndvi-timeseries}
\end{figure}

\glsunset{vae}

Advances in hardware and algorithms are encouraging signs that quantum computers will have useful applications \cite{preskill2018quantum, arute2019quantum}, though hard challenges remain.
One of these challenges is finding early target applications.
\Gls{qml} is being explored as a tool to demonstrate quantum advantage \cite{biamonte2017quantum}, a situation where a quantum computer is preferred over a classical one. 
For example, the \gls{qvae} \cite{khoshaman2018quantum,vinci2019path} integrates samples from a quantum annealer into the prior of a discrete \gls{vae} \cite{rolfe_discrete_2016, kingma_auto-encoding_2013}.
Quantum annealers can be thought of as analog quantum devices that can physically simulate a \gls{qbm} \cite{amin2018quantum}. 
Quantum and classical \gls{bm} are able to model powerful and flexible distributions. 
Training such model, however, is known to be difficult and slow classically \cite{long2010restricted}. 
The use of quantum annealers as samplers could result in more robust and scalable sampling, and allow to train \glspl{qbm} to be used in models of practical impact. 
In this work, we investigate the real-world applicability of a \gls{qvae} to the problem of similarity search in large-scale high-dimensional datasets.
Though, it should be clear that we explore the feasibility and are neither achieving nor trying to achieve state-of-the-art performance.

Similarity search is an important problem for many applications such as computer vision and recommender systems and has been extensively studied in the literature.
However, few have focused on scaling to billions of data points with thousands of dimensions \cite{li_approximate_2016}.
The methodology towards scaling to billions of data points and to very high-dimensional space differs.
While research on scaling to billion-scale datasets focused on efficiently utilizing current hardware and accelerators \cite{camerra2014beyond, johnson_billion-scale_2017, yagoubi_radiussketch:_2017}, recent work on very high-dimensional data suggests the use of deep neural networks as dimensionality reduction tools \cite{do_learning_2016, srivastava_unsupervised_2015, carreira-perpinan_hashing_2015, cao_correlation_2016}.

With increasing size and dimensionality of datasets, this issue is gaining importance.
Massive datasets push classical algorithms to their limits.
An area especially affected by this is the Earth science community where datasets grow exponentially with increasing spatial and temporal resolution.
NASA, for instance, operates hundreds of satellites to collect continuous data of Earth's processes to study land-cover change, atmospheric systems, and many others~\cite{board2007earth}. 
But, the ever-growing size of these datasets poses challenges to scientists studying the Earth system. 

We present an \gls{anns} approach which uses a \gls{qvae} to embed data into an efficiently searchable binary Hamming space.
The binary nature of the \gls{qvae} latent space removes the necessity for any additional discretization and enables end-to-end training.
Further, we evaluate the approximation quality of the Hamming distance in the compressed space, the effect of the quantum prior, the resulting speedup and the memory footprint using the \gls{modis} dataset.
An example of our \gls{anns} is shown in Figure~\ref{fig:ndvi-timeseries} where a reference point with all candidates with Hamming distance $\leq 1$ are shown in Figure~\ref{fig:ndvi:map} and a comparison with a random candidate is shown in Figure~\ref{fig:ndvi:comp}.

To the best of our knowledge, this is the first work that explores the applicability of \gls{qml} to similarity search in large-scale high-dimensional datasets.
Further, we demonstrate the usefulness of a quantum hardware specific feature, the transverse field, in shaping the latent distribution to improve real-world performance and present a framework to index hundreds of millions of data points with minimal space requirements.

The remainder of this paper is structured as follows. In Section~\ref{sec:related_work} we provide the necessary background and position our work in the context of recent research.
We present our method in Section~\ref{sec:method} and evaluate it in Section~\ref{sec:experiments}.
The result are discussed in Section~\ref{sec:discussion} and we conclude in Section~\ref{sec:conclusion}.

\section{Related Work}\label{sec:related_work}
In the following, we present related work and background.
We start with the motivation for high-dimensional similarity search in \acrlong{es} in Section~\ref{sec:related:es}.
Next, we examine existing \gls{anns} methods and distinguish our approach from existing work in Section~\ref{sec:related:ann}.
Section~\ref{sec:related:vae} provides background on \glspl{vae} and the transition to a \gls{qvae}, i.e., a \glspl{vae} with a \gls{qbm} prior, which in turn is explained in Section~\ref{sec:related:qbm} in addition to \glspl{bm} in general.

\subsection{Earth Science}\label{sec:related:es}
Satellite observations, physical model outputs, and ground-based systems collect spatio-temporal data that are used to understand changes in the Earth system. 
Analysis of these datasets have allowed scientists to uncover changes in interannual variability, anomalies~\cite{wright2002automated}, location similarity, and seasonal phenomenology \cite{zhang2003monitoring}. 
While many currently available geospatial tools provide easy access to spatial processing, tools for large-scale similarity of high-resolution datasets are not as well developed.
As data in the Earth sciences continues to grow and available applications broaden, computational efficiency will gain importance. 

Steady improvements to sensing hardware and high-performance computing enable higher spatial, temporal, and spectral resolution remote sensing data and physical model outputs. 
However, improved resolution causes an exponential increase in data storage requirements leading to large-scale analysis challenges. 
For example, as we will use in this study, the \acrfull{modis} dataset \cite{remer2005modis} from NASA's Terra and Aqua satellites provides 17 years of daily observations of the entire Earth's surface at a spatial resolution of 250 meters over 36 spectral bands at the petabyte scale.
After data collection, higher-level products are then generated from raw sensor output using physics-based pre-processing for specific applications such as land-surface temperature, snow-cover, and fire. 
For instance, the \gls{modis} Vegetation dataset is widely used to monitor seasonal variations in phenology at a global scale \cite{zhang2003monitoring}. 
As satellite and climate datasets continue to improve, high-dimensional data analysis will continue to be a burden to researchers. 

In recent years data mining tools have been commonly used to study large-scale Earth science datasets for land-cover change detection~\cite{boriah2008land}, climate networks~\cite{steinhaeuser2012multivariate}, and many others~\cite{kamath2001mining}. 
Climate networks, for example, are used to locate climatic teleconnections by quantifying dependencies between all locations with the dataset~\cite{zhou2015teleconnection}. 
For high-resolution data, exhaustive approaches to generate these complex networks become infeasible without an efficient similarity search. 
Here, we explore the applicability of \glspl{qvae} to index multi-variate time-series of land cover change using the \gls{modis} dataset.
\subsection{Approximate Nearest Neighbor Search}\label{sec:related:ann}
Finding nearest neighbors is of importance in a variety of areas, e.g., databases, computer vision or recommender systems.
However, performing an exact nearest neighbor search is computationally expensive for large high-dimensional datasets with $\dd\gg\log{n}$, where $n$ is the size of the dataset and $\dd$ its dimensionality, due to the curse of dimensionality ~\cite{indyk_approximate_1998}.

In many applications it is sufficient to find near objects instead of nearest, motivating approximate methods.
Generally, \gls{anns} algorithms rely on pre-computed indices whose initial build may take significant time but reduce the query time by orders of magnitude.
A variety of approaches have been proposed but most can be categorized into hashing or graph-based methods \cite{li_approximate_2016}.
Since our method falls into the first category we will elaborate more on hashing-based approaches here.
The most prominent example is \gls{lsh} \cite{gionis_similarity_1999}.
It uses $L$ hash tables, dictionaries that map from a hash to all items with this hash, as index.
As hash functions \gls{lsh} relies on ensembles of similarity preserving hash functions.
At query time, items that agree in at least one hash are returned as candidates and later on refined by linear search. 
Although \gls{lsh} performs well on low-dimensional datasets, high-dimensional datasets usually require prior dimensionality reduction.

Research into using deep neural networks, especially autoencoders, extends hashing based approaches by learning the hash function.
While this approach showed promising results for high-dimensional data it comes with some disadvantages:
The discrete space of a hash function yields a non-smooth optimization problem which has often been solved by using complex alternating optimization algorithms ~\cite{do_learning_2016, carreira-perpinan_hashing_2015} or by relaxing to a continuous space ~\cite{srivastava_unsupervised_2015}.
Further, labeled data is rarely available for similarity search, thus most methods must rely on unsupervised learning \cite{do_learning_2016, srivastava_unsupervised_2015, carreira-perpinan_hashing_2015}.
But supervised ~\cite{cao_correlation_2016} and semi-supervised methods exist ~\cite{do_learning_2016}.

In this work, we use a \gls{qvae} to learn hash functions in a Hamming space. 
This setup yields several advantages: It relies on the \gls{vae} framework to perform fully scalable, unsupervised analysis of high-dimensional data. It exploits a generalization to discrete latent-variable \gls{vae} \cite{rolfe_discrete_2016} to enable end-to-end training with gradient descent while maintaining a binary latent space at inference. Furthermore, \glspl{vae} are probabilistic autoencoders, so our method ensures similarity between neighboring hashes enforcing the relevance of the Hamming distance. Finally, as we discuss in further detail in Section~\ref{sec:related:qbm}, the integration of a \gls{qbm} prior allows continuous control over the distribution of the latent variables by adjusting the tunneling term (transverse field) in the \gls{qbm}. 

\subsection{Quantum Variational Autoencoder}\label{sec:related:vae}

A \gls{vae} is a generative model with latent variables $\mathbf{z}\sim p_\theta(\mathbf{z})$ which consist of two neural networks, an encoder $q_\phi(\mathbf{z}|\mathbf{x})$ and a decoder $p_\theta(\mathbf{x}|\mathbf{z})$ parametrized by $\phi$ and $\theta$, respectively.
It avoids the problem of computing the true posterior $p(\mathbf{z}|\mathbf{x})$ by using a variational approximation $q_\phi(\mathbf{z}|\mathbf{x})$ \cite{hoffman_stochastic_2013}, i.e., it is trained by maximizing a tractable \gls{elbo} of the log likelihood $\mathbb{E}_{p(\mathbf{x})}[\log{p_\theta(\mathbf{x})}]$:
\begin{align}
    \mathcal{L}(x) &= 
        \mathbb{E}_{p(\mathbf{x})}f[
            \mathbb{E}_{q_\phi(\mathbf{z}|\mathbf{x})}[\log{p_\theta(\mathbf{x}|\mathbf{z})}
        ]\label{eq:vae_loss}\\
        &-D_{KL}(q_\phi(\mathbf{z}|\mathbf{x})||p_\theta(\mathbf{z}))]\nonumber,\\
    \mathcal{L}(\mathbf{x}) &\leq
        \mathbb{E}_{p(\mathbf{x})}[\log{p_\theta(\mathbf{x})}].
\end{align}
The first term in Equation~\eqref{eq:vae_loss} is known as the reconstruction loss and $D_{KL}(q||p)$ is the \gls{kl} divergence between the distributions $q$ and $p$.

The reparametrization trick \cite{kingma_auto-encoding_2013} stabilizes the gradients by introducing an auxiliary random variable $\rho$ and the reparametrization function $g$.
The approximate posterior $q_\phi(\mathbf{z}|\mathbf{x})$ is then rewritten as $q_\phi(\mathbf{z}|\mathbf{x},\rho)=q_\phi(g(\zeta, \rho)|\mathbf{x})$ with $\zeta\in\mathbb{R}$ being `logits' outputted by the encoder.
This allows us to reformulate the \gls{kl} divergence:
\begin{align}
    D_{KL}(q_\phi||p_\theta) &=
        \mathbb{E}_{q_\phi}[{\log q_\phi}] -
        \mathbb{E}_{q_\phi}[\log p_\theta] \label{eq:kl},\\
    \mathbb{E}_{q_\phi}[{\log q_\phi}] &= 
        \mathbb{E}_{p(\rho)}[\log q_\phi(\mathbf{z}|\mathbf{x},\rho)]\label{eq:reparam},
\end{align}
where we used $q_\phi$ and $p_\theta$ in place of $q_\phi(\mathbf{z}|\mathbf{x})$ and $p_\theta(\mathbf{z})$ to avoid clutter.
Computing $\mathbb{E}_{q_\phi}[\log p_\theta]$ depends on the prior distribution while $\mathbb{E}_{q_\phi}[\log q_\phi]$ depends on the reparametrization.
Since our method requires a discrete latent space, we will focus on \glspl{bm} as priors and binary latent spaces $\mathbf{z}\in\{0,1\}^d$ from here. 
The generalization of the reparametrization trick to discrete variables is performed following Reference \cite{khoshaman2018gumbolt}, which introduces a smoothed continuous approximation of the latent discrete variables as
\begin{align}
    g(\zeta,\rho)&=\sigma((\zeta+\log({\rho}) - \log(1-\rho)) * \alpha), \label{eq:reptrick}\\
    \text{with } \sigma(t)&=\frac{1}{1+\exp(-t)},
    \rho\sim \mathcal{U}(0,1)\nonumber
\end{align}
and $\alpha\in\mathbb{R}_+$ regulating the `strength' of the discretization.
To get a deterministic hash function, we decided to map to the state with the highest likelihood $\mathbf{z} = \frac{\text{sign}(\zeta)+1}{2}$ instead of sampling and discretizing during inference.
Using this Bernoulli reparametrization, the closed form solution for Equation~\eqref{eq:reparam} is the element-wise cross entropy~\cite{khoshaman2018quantum}:
\begin{align}
    \mathbb{E}_{q_\phi}[{\log q_\phi}] &= 
        \mathbb{E}_{q_\phi}\left[
            \sum_{i=1}^d z_i\log(\sigma(\zeta_i))+(1-z_i)\log(1-\sigma(\zeta_i))
        \right].
\end{align}
If $p_\theta$ is distributed by an \gls{bm}, the gradients of $\mathbb{E}_{q_\phi}[\log p_\theta]$ can be computed the same way as training the \gls{bm} with $q_\phi$ as target distribution, as described in Section \ref{sec:related:qbm}, with the gradients propagating through the samples of the positive phase to the encoder ~\cite{rolfe_discrete_2016}.

Replacing the \gls{bm} prior with a \gls{qbm} yields the \gls{qvae}.
The \gls{qbm} offers an additional parameter, the transverse field which enables continuous control over the prior distribution, concretely, by increasing the transverse field, the prior loses expressive power.
Previous work~\cite{hoffman_elbo_2016, so_nderby_ladder_2016, casale_gaussian_2018} has shown that the expressive power of the latent prior is linked to the number of decoupled bits in the latent distribution, i.e., decreasing the expressiveness of the prior narrows the latent distribution.
In the discrete case, this causes the encoder to map to fewer distinct bit strings.
Further discussion on the transverse field can be found in Section \ref{sec:related:qbm}.

\subsection{Quantum Boltzmann Machines}\label{sec:related:qbm}
\begin{figure}
    \centering
    \begin{subfigure}{.49\linewidth}
        \includegraphics[width=\linewidth]{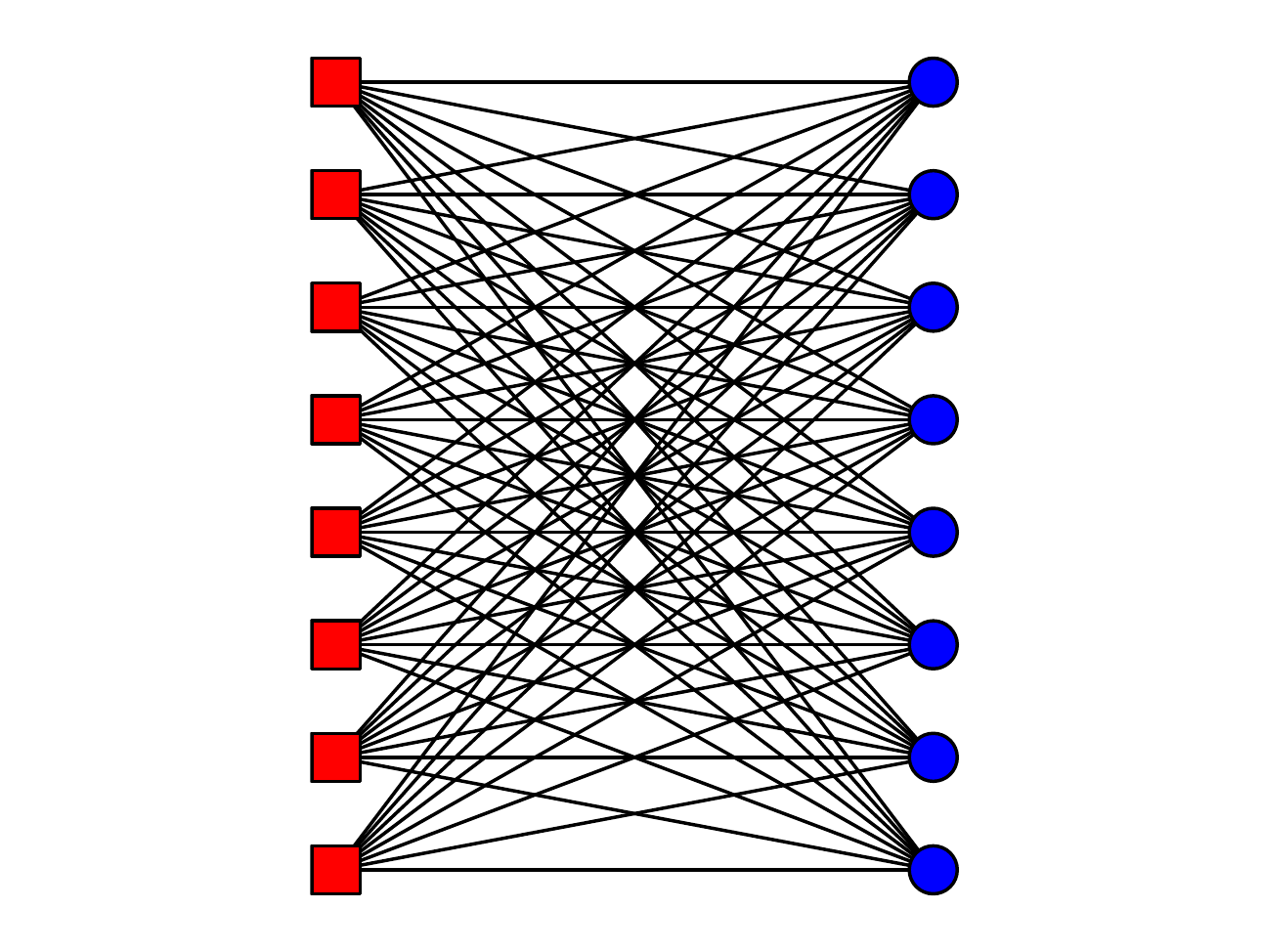}
        \caption{Logical graph}
        \label{fig:rbm}
    \end{subfigure}
    \begin{subfigure}{.49\linewidth}
        \includegraphics[width=\linewidth]{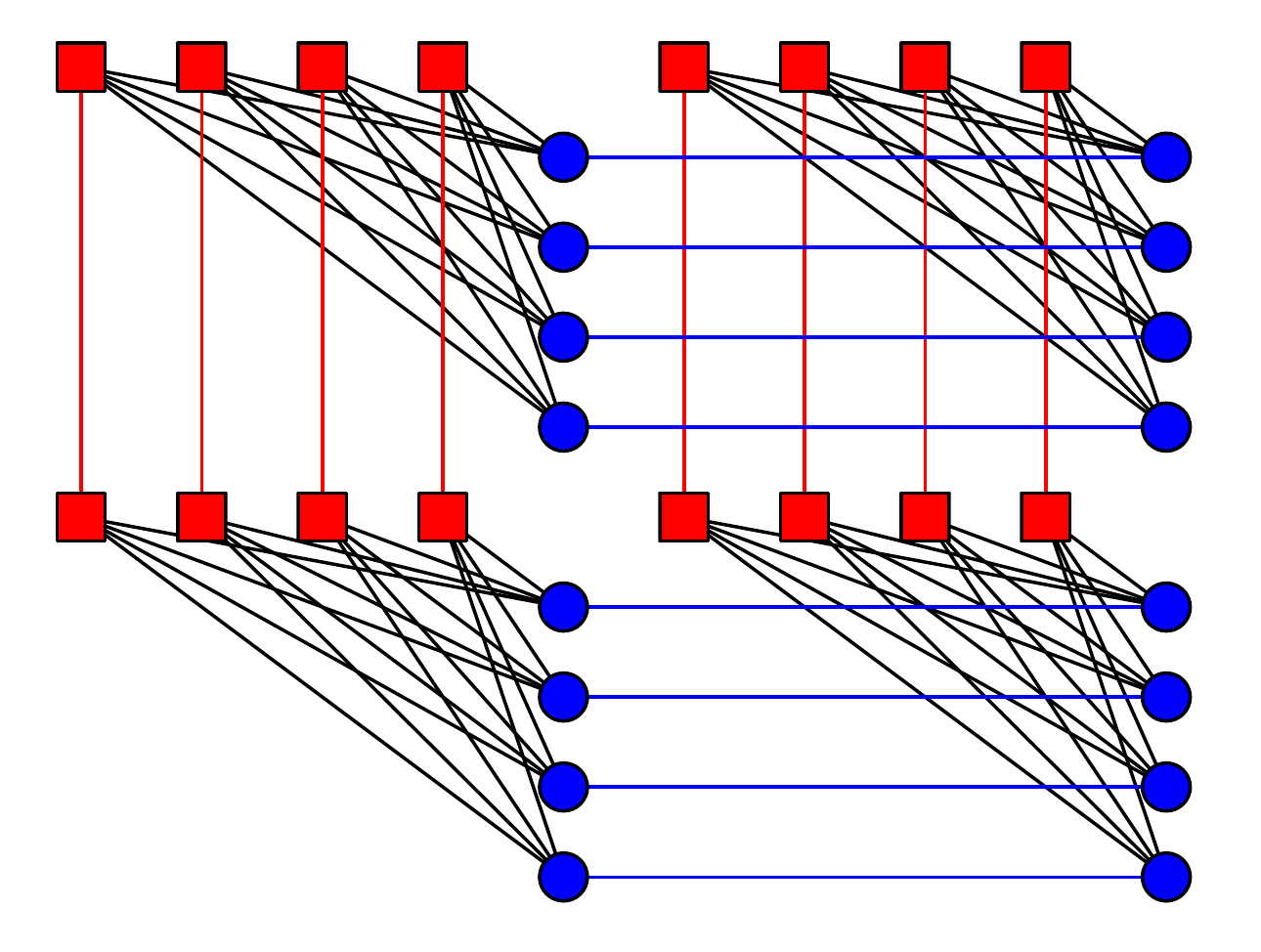}
        \caption{Hardware embedding}
        \label{fig:dwave}
    \end{subfigure}
    \caption{
        Representations of a $8\times 8$ \acrshort{rbm}. The red squares represent the left nodes and the blue spheres represent the right nodes.
        Figure~\ref{fig:rbm} shows the logical graph and Figure~\ref{fig:dwave} a hardware embedding where chains of qubits form single logical nodes.
    }
    \label{fig:embedded_rbm}
\end{figure}

A \gls{bm} is an energy-based graphical model composed of stochastic nodes, with weighted connections between and biases applied to the nodes, i.e., a weighted graph $\mathcal{G} =(\mathcal{V}, f, g)$ with nodes $\mathcal{V}$, node weight function $f:\mathcal{V}\mapsto\mathbb{R}$ and edge weight function $g:~\mathcal{V}~\times~\mathcal{V}~\mapsto~\mathbb{R}$ describes a \gls{bm} over the states $\mathcal{Z}=\{0, 1\}^{|\mathcal{V}|}$ with parameters $\theta = \{\boldsymbol{\omega}, \boldsymbol{b}\}$ where $\omega_{ij}=g(v_i,v_j)$ and $b_i=f(v_i)$.
We refer to this graph as the logical graph.
The energy for a state $\mathbf{z}\in\mathcal{Z}$ is given by 
\begin{equation}
    E_{\theta}(\mathbf{z}) = -\sum\limits_{z_i \in \mathbf{z}} b_{i} z_i - \sum\limits_{(z_i,z_j) \in \mathbf{z} \times \mathbf{z}} \omega_{ij} z_i z_j
    \label{eq:boltzmann_machine}
\end{equation}
and the \gls{pmf} of the state distribution is
\begin{equation}
    p_{\theta}(\mathbf{z}) = \frac{e^{-\beta E_{\theta} (\mathbf{z})}}{Z}, \text{with }
    Z = \sum_{\mathbf{z}\in\mathcal{Z}}e^{-\beta E_{\theta}(\mathbf{z})}
    \label{eq:boltzmann_distribution}
\end{equation}
being an intractable normalization constant and $\beta$ a parameter recognized by physicists as the inverse temperature.

A \gls{bm} is usually trained via \gls{cd} \cite{hinton_training_2002,tieleman2009using}.
Given a target distribution $q$, and a \gls{bm} parametrized by $\theta$, the gradients can be computed as
\begin{align}
    \mathbb{E}[\partial \theta] &=
    \underbrace{
        \mathbb{E}_{z\sim q}[\beta\partial E_{\theta}(\mathbf{z})]
    }_\text{positive phase}
    -
    \underbrace{
        \mathbb{E}_{z \sim p_{\theta}}[\beta \partial  E_{\theta}(\mathbf{z})]
    }_\text{negative phase}.
    \label{eq:rbmprior}
\end{align}
The positive phase are the gradients of the energy for samples from the target distribution while the negative phase is computed with samples from the \gls{pmf} in Equation~\eqref{eq:boltzmann_distribution}.
Since it is hard to sample and compute probabilities on fully-connected \glspl{bm}~\cite{koller2007graphical}, \glspl{rbm} are commonly used as these can be more efficiently sampled using block-Gibbs sampling.
An \gls{rbm} is a symmetric complete bipartite \gls{bm} as shown in Figure~\ref{fig:rbm}.

The next paragraph describes the sampling from a quantum annealer and related background. However, the quantum annealer may be treated as a block-box sampler, returning samples $\mathbf{z}$ given parameters $\theta$ and a transverse field, and the paragraph be skipped.

\gls{qa}, an optimization algorithm exploiting quantum phenomena, has been proposed for sampling from complex Boltzmann-like distributions.
\gls{qa} has been demonstrated to solve a range of instances of optimization problems \cite{rieffel2019ans}, though it remains unclear what speedup \gls{qa} can provide; even defining and detecting speedup, especially in small and noisy hardware implementations is challenging \cite{ronnow2014defining, katzgraber2014glassy}.
To sample a \gls{qbm} using \gls{qa}, the framework outlined in Equation~\eqref{eq:boltzmann_machine} has to be mapped to an Ising model, a quantum system represented by the Hamiltonian
\begin{equation}
    \hat{H}_{p} = - \sum\limits_{\hat{\sigma}_z^{(i)} \in \mathcal{Z}} h_i \hat{\sigma}_z^{(i)} - \sum\limits_{(\hat{\sigma}_z^{(i)},\hat{\sigma}_z^{(j)}) \in \mathcal{Z}\times\mathcal{Z}}  J_{ij} \hat{\sigma}_z^{(i)} \hat{\sigma}_z^{(j)}
    \label{eq:ising}
\end{equation}
where the nodes $z_i$ have been replaced by the quantum operators $\hat{\sigma}_z^{(i)}$, which return eigenvalues, $\lambda_i$, in the set $\{-1, 1\}$ when applied to the state of variable $z_i$, corresponding to 0 and 1, respectively. The possible states in the `classical' and Ising representations are related by $\lambda_i = 2z_i - 1$. Parameters $b_i$ and $\omega_{ij}$ are replaced with the Ising model parameters $h_i$ and $J_{ij}$ which are conceptually equivalent. In the hardware, these parameters are referred to as the flux bias and the coupling strength, respectively. 
The full model describing the dynamics of the quantum annealer, equivalent to the time-dependent transverse field Ising model, is \cite{amin2018quantum}
\begin{equation}
    \hat{H}(t) = A(t) \hat{H}_\perp + B(t) \hat{H}_{p}
    \label{eq:anneal}
\end{equation}
where the transverse field term $H_\perp$ is
\begin{equation}
    \hat{H}_\perp = \sum\limits_{\hat{\sigma}^{(i)}_x \in V} \hat{\sigma}^{(i)}_x,
    \label{eq:transverse}
\end{equation}
with $\hat{\sigma}_x$ being quantum operators on the complex-valued Hilbert space $V=\mathbb{C}^{2^{|\mathcal{V}|} \times 2^{|\mathcal{V}|}}$, $A(t)$ and $B(t)$ being monotonic functions and $t_\mathrm{max}$ being the total annealing time \cite{biswas2017nasa}. Generally, at the start of an anneal, $A(0)\approx1$ and $B(0)\approx0$. 
$A(t)$ decreases and $B(t)$ increases monotonically with $t$ until, at the end of the anneal, $A(t_\mathrm{max})\approx0$ and $B(t_\mathrm{max}) \approx 1$. When $A(t) > 0$, Equation~\eqref{eq:anneal} contains terms that describe quantum tunneling effects between various units (the transverse field $H_\perp$) that are not present in the classical model.

At a high level, increasing the transverse fields increases the amount of quantum fluctuations present in the model.
These quantum fluctuations move the classical Boltzmann distribution towards a binomial distribution with $p~=~0.5$.
This reduction impedes the ability to capture correlations between bits and causes the distribution to lose expressive power.

Lastly, the connectivity of the \gls{bm} may not be reflected in the connectivity between qubits in the annealer. If that is the case, one needs to embed the logical graph into a hardware-specific one.
This can be done by a 1-to-many mapping, i.e., one variable $z_i$ is represented by multiple qubits.
These qubits are arranged in a `chain' by setting the coupling strength $J_{ij}$ between these qubits to a large value to encourage them to take the same value.
Full details of the embedding process used in this model can be found in Reference~\cite{adachi2015application}.
An example of an embedded graph with maximum degree 6 can be found in Figure~\ref{fig:dwave} where chains of red squares and chains of blue circles represent the \gls{rbm} nodes.

\section{Method}\label{sec:method}
\begin{figure*}
    \centering
    \includegraphics[width=\textwidth]{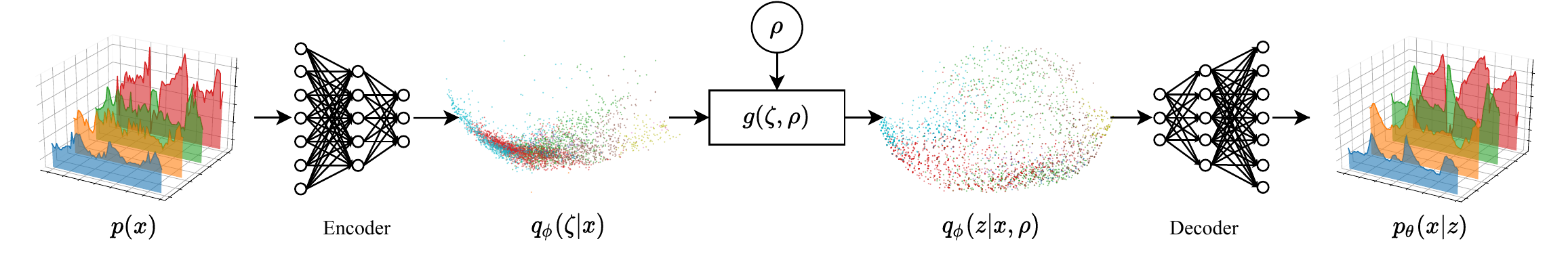}
    \caption{
        \gls{qvae} pipeline: The sequence from left to right represents the forward pass of our model where $p(\mathbf{x})$ is the data distribution, $q_\phi(\zeta|\mathbf{x})$ the distribution of `logits', $\rho\sim\mathcal{U}(0,1)$ is a uniformly distributed random variable, $g(\zeta,\rho)$ the reparametrization function, $q_\phi(\mathbf{z}|\mathbf{x},\rho)$ the discrete latent distribution and $p_\theta(\mathbf{x}|\mathbf{z})$ distribution of the reconstruction.
    }
    \label{fig:architecture}
\end{figure*}

This section is dedicated to describing our approach on \gls{qvae}-based \gls{anns}. Section \ref{sec:qvae} covers the implementation of our \gls{qvae} and Section \ref{sec:method:ann} focuses on how to perform \gls{anns}.

\subsection{Model}\label{sec:qvae}
The model architecture is depicted in Figure \ref{fig:architecture}.
The network represents a \gls{vae} as described in Section \ref{sec:related:vae}.
As encoder it uses two fully-connected layers followed by a hierarchical posterior from Reference \cite{rolfe_discrete_2016}.
The hierarchical posterior allows the modeling of more complex posterior distributions and thus optimize a tighter \gls{elbo}.
To avoid overfitting, i.e.,  encoding the data distribution in the decoder's parameters, the decoder network is simpler by only featuring two full-connected layers.
All layers use the \gls{relu} activation and we use the reparametrization of discrete latent variables from Equation~\eqref{eq:reptrick} with $\alpha=7$ \cite{khoshaman2018gumbolt, vinci2019path}.

We scale our data to be zero-centered with unit variance and choose a multivariate normal distribution with unit variance as reconstruction target.
Therefore, our reconstruction loss, Equation~\eqref{eq:vae_loss}, is the mean squared error defined by
\begin{align}
    \mathbb{E}_{q_\phi(\mathbf{z}|\mathbf{x})}[\log{p_\theta(\mathbf{x}|\mathbf{z})}] =
    \frac{1}{2}\mathbb{E}_{p(\mathbf{x})}\left[||\mathbf{x}-\Phi_{\phi,\theta}(\mathbf{x})||_2^2\right] \label{eq:mse}
\end{align}
where $\Phi_{\phi,\theta}$ is the forward pass of our model, i.e., mapping to the latent space and back to the original space.

We compare an \gls{rbm}, a classically simulated \gls{qbm} and a \gls{qbm} trained with samples drawn from a quantum annealer.
Parameter gradients are computed by \gls{cd}, Equation~\eqref{eq:rbmprior}, for all priors.

The quantum annealer which we used in the experiments, a D-Wave 2000Q, has 2048 qubits, each of degree 6.
Since any \gls{rbm} with more than 6 nodes per side is not native to the hardware, we must embed the logical graph by the methods outlined in Section~\ref{sec:related:qbm} with a chain coupling strength of -1, the maximum allowed by the D-Wave 2000Q. 
At the end of an anneal, the value of a logical variable is determined by a majority vote.

The quantum annealer samples at some unknown effective temperature $\beta^*$, analogous to the `temperature' $\beta$ of the classical \gls{rbm}, Equation~\eqref{eq:boltzmann_distribution}. 
In the case of a standalone \gls{rbm}, changing $\beta$ is equivalent to scaling the learning rate during training, Equation~\eqref{eq:rbmprior}. 
But, this does not apply for an \gls{rbm} placed as prior of a \gls{vae}, due to the gradients of the encoder also propagating through the positive phase \cite{vinci2019path}. We estimate this parameter throughout the training with an auxiliary \gls{rbm}.
We define the auxiliary \gls{rbm} with the same parameters as the \gls{qbm} and add the effective temperature $\beta_\mathrm{eff}$ as the only free variable, i.e., the \gls{pmf} of the auxiliary \gls{rbm} is given by $
    p_{\theta}(\mathbf{x}) = \frac{e^{-\beta_{\mathrm{eff}} E_{\theta}(\mathbf{x})}}{Z_{\theta}}$.
We estimate the effective temperature by optimizing w.r.t. to $\beta_\mathrm{eff}$ using \gls{cd} as in Equation~\eqref{eq:rbmprior}.
The negative samples are drawn via Gibbs sampling while the positive samples are from the quantum annealer. When sampling the auxiliary \gls{rbm}  we use the effective parameters $b_\mathrm{eff}~=~ b*\beta_\mathrm{eff}$ and  $\omega_\mathrm{eff}=\omega*\beta_\mathrm{eff}$. 
While this technique is not scalable to large \glspl{bm}, it is likely that the effective temperature  $\beta^*$ at which quantum annealers sample can be fixed and stabilized with improved control of the annealing schedule.
For the classical and simulated model, we fix $\beta=1$.

Finally, parameters on the D-Wave are restricted to  $-1\leq\omega_i\leq1$.
To maximize similarity between the auxiliary \gls{rbm} and the quantum annealer and to limit the magnitude of these parameters, we apply a L2 regularization to the parameters of the \gls{qbm} and project them back on to the unit L1 box after every update step.
All updates for the model are computed using Adam \cite{kingma_adam_2017}.

\subsection{Search Algorithm}\label{sec:method:ann}
Here we describe the search algorithm used and provide complexity bounds on all operations. Throughout this section, $n$ denotes the size of the dataset and $m$ the number of unique bit strings.
For simplicity, we assume the dimensionality of the original space $\dd$ and of the latent space $\dl$ as fixed since $\dl<\dd\ll m\leq n$.
It should be noted that while we use the Euclidean distance here, our approach is also applicable for any other distance metric given Equation~\eqref{eq:mse} is adjusted accordingly.

After training, we use the encoder of our \gls{qvae} to index the data.
As discussed in Section \ref{sec:related:vae}, we use $\mathbf{z}=\frac{\mathrm{sign}(\zeta) + 1}{2}$ during encoding instead of the reparametrization, Equation~\eqref{eq:reptrick}.
Due to the finite number of states in our latent space, the encoding of similar objects will map to the same bit string.
We then construct an inverted index, a hash map that maps from a bit string to all data points which have been mapped to that bit string.
Since every item is only processed once and independently this operation is in $O(n + m)$ and the original data can be removed from main memory after processing reducing memory usage.

Given a query item and a number of requested neighbors $k$, we first encode the query item with the encoder, which is a constant time operation, and sort all occupied bit strings by their Hamming distance to the query embedding.
The computation of the Hamming distance and search are both in $O(m)$.
While naively sorting requires $O(m\log m)$ operations, we can reduce this to $O(m)$ by taking advantage of the discrete values of the Hamming distance and RadixSort.
We also explored the use of iterative Branch-and-Bound algorithms but found it be slower than RadixSort when thousands of bit strings have to be iterated in a sparse index with $m\ll 2^\dl$.
Next, the items are iterated in order of their Hamming distance and compared to the query item in terms of Euclidean distance in the original space.
We stop the iteration after a fixed number of comparisons $c_{\max}$ with $k~\leq~c_{\max}~\ll~n$ to keep a fixed search time independent of the query item but other stopping criteria are also possible, e.g. by Hamming distance or by the number of visited hashes.
After stopping, the $k$ best results are returned which can be done in $O(c_{\max})$ by partitioning.
The complete runtime of a single search pass is therefore in $O(m+c_{\max})$.

The memory usage scales linear with the number of objects and is constant with respect to the original dimensionality of the data.
In fact, we only need to store every bit string once and a single 8 byte integer per object, so the total space requirement is $O(m\dl+n)$, as $m\leq n$ we can simplify this to $O(n)$ for a fixed $\dl$.

Although smaller $m$ decrease the search time and memory footprint, increasing the number of distinct bit strings yields a higher resolution space and thus better search results.
To control the number of occupied bit strings, one could adjust the latent dimension $\dl \in \mathbb{N}_+$ but since the maximum number of distinct bit strings $m_{\max}$ scales exponentially with $\dl$ this only allows for coarse control.
We propose to use the tunable quantum effects of a \gls{qbm}, the transverse field defined in Equation~\eqref{eq:transverse}, to gain continuous control over the latent distribution. 
As discussed in Section \ref{sec:related:qbm}, the transverse field affects the expressiveness of our prior and allows us to narrow the latent distribution while keeping the network's ability to learn how many states should be occupied.

\section{Experiments}\label{sec:experiments}
We conducted a series of experiments on the \gls{modis} dataset to evaluate our approach.
The exact dataset definitions can be found in Section~\ref{sec:dataset}.
As first experiment, in Section~\ref{sec:experiments:latent}, we trained a model with an \gls{rbm} prior and performed \gls{anns} to verify that the Hamming distance in the compressed space functions as a proxy to the Euclidean distance in the original space.
We then classically simulated sampling from the quantum distribution to evaluate the effect of the transverse field parameter on the latent space and search speed in Section~\ref{sec:experiments:transverse}.
Next, we compared models trained via quantum simulation, classical sampling and hardware sampling in terms of speedup by recall in Section~\ref{sec:experiments:speedup}.
For this experiment, we did not include comparisons to other methods as the goal of this work and the associated implementation has not been to push the limits in terms of speed or accuracy. We discuss this further in Section~\ref{sec:discussion}.
Lastly, we investigated the memory consumption and show how our method scales to half a billion data points in Section~\ref{sec:experiments:memory}.

\newcommand{\tqvae}{{t_{\mathrm{QVAE}}}}
\newcommand{\tlinear}{{t_{\mathrm{linear}}}}
\newcommand{\Rqvaek}{{R_{\mathrm{QVAE}}^k}}
\newcommand{\Rlineark}{{R_{\mathrm{linear}}^k}}
We compare all of our results to a baseline linear search algorithm and report time improvements in terms of speedup $\frac{\tlinear}{\tqvae}$ where $\tqvae$ and $\tlinear$ are the averaged elapsed wall times a query takes to complete for our QVAE approach and linear search, respectively.
The quality of the search is assessed by the recall  $\frac{|\Rqvaek\cap \Rlineark|}{k}$, where $\Rqvaek$ and $\Rlineark$ are the first $k$ results from our QVAE search and linear search.

We executed all experiments on a computer with 36 CPU cores and 384GB of RAM.
When reporting timings or speedup we loaded the complete dataset into the main memory to remove hard drive access from wall time measurements.
We used a D-Wave 2000Q quantum annealer and simulated the sampling with Quantum Monte Carlo~\cite{gull2011continuous}.
The dimensions of the two hidden layers found in the encoder and decoder have been fixed to 128 and 64.
We fixed the latent size to 64 by picking the latent size out of 32, 64 and 128 with highest speedup on the \gls{rbm} model at a recall of 0.8.

\subsection{Dataset}\label{sec:dataset}
The proposed approach has been evaluated for the problem of detecting similarities of land-cover vegetation over multiple years using the \gls{modis} dataset on the Terra satellite ~\cite{justice1998moderate}.
Terra is a polar orbiting satellite that covers the entire globe once per day at approximately 10:30am capturing 36 spectral bands at a 250m-2km spatial resolution. 
In this work, we used the \gls{modis} Terra 16-day 500 meter vegetation indices (MOD13A1.006) from January 2016 to December 2018~\cite{solano2010modis}, 69 time-steps.
Each \gls{modis} tile covers a region of 1200$km$ by 1200$km$ with 2400 by 2400 pixels.
Six variables were selected: Normalized Vegetation Index (NDVI), Enhanced Vegetation Index (EVI), red reflectance (0.645$\mu m$), near infra-red reflectance (0.858$\mu m$), blue reflectance (0.469$\mu m$), and  middle infra-red reflectance (2.13$\mu m$).
We ignored all pixels with missing data.
If not otherwise stated, we used the tiles from the rectangular region covering North America, h08v04 to h12v06.
After filtering, 45,386,870 data points remained with 414 dimensions.

\subsection{Embedded Proximity}\label{sec:experiments:latent}
\begin{figure}
    \centering
    \includegraphics[width=\linewidth]{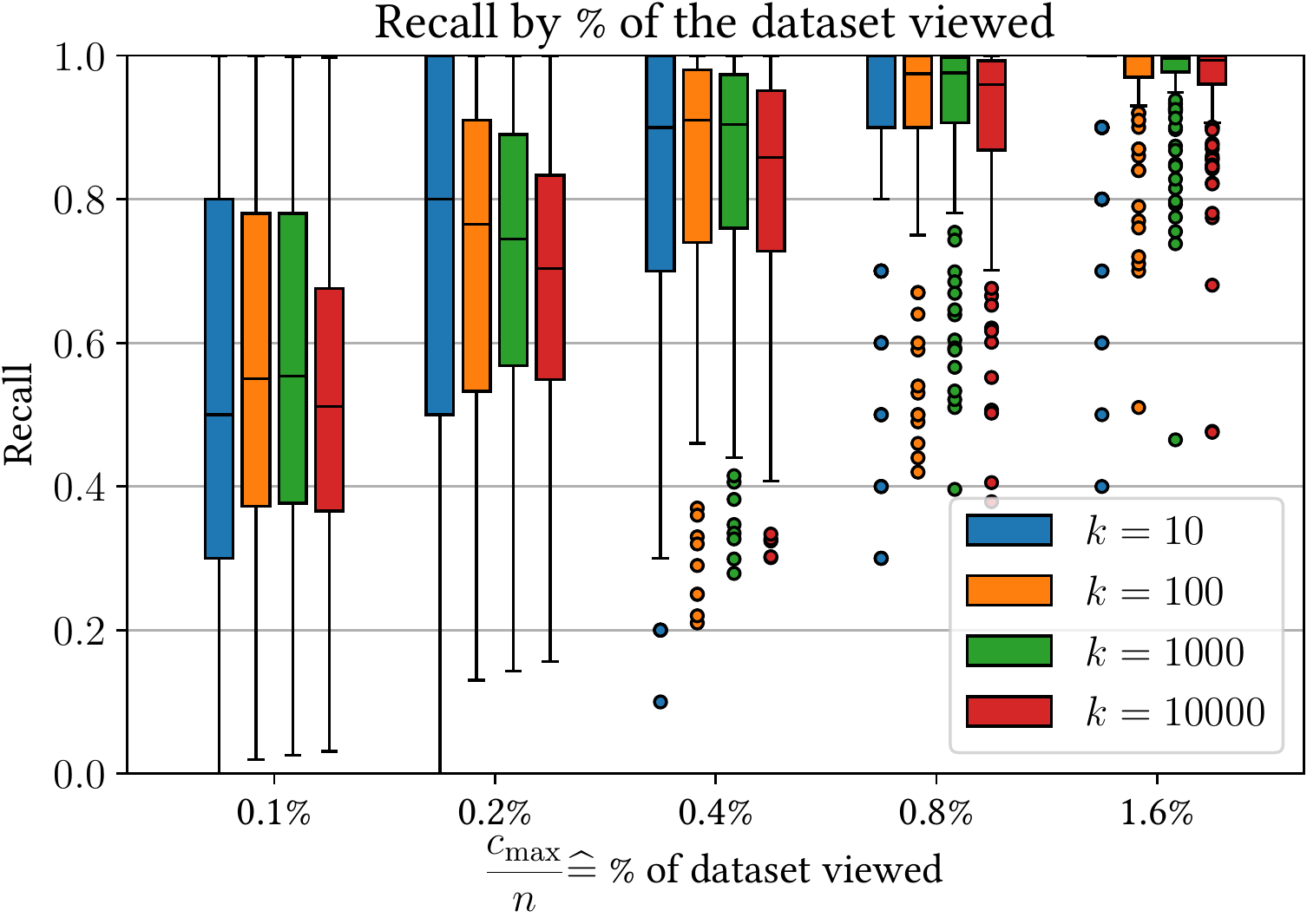}
    \caption{
        Recall value by the \% of the dataset has been plotted for different $k$. Large recall values are preferred. The box extends to the lower and upper quartile values of the data, the black line within represents the median, the whiskers show the range of the data and outliers are marked as circles.
    }
    \label{fig:recall_by_hamming}
\end{figure}
The quality of approximation of the Hamming distance in the compressed space to the Euclidean distance in the original space is crucial to the functionality of our search.
To validate this assumption, we trained a model with an \gls{rbm} prior and performed $k$-\gls{anns}, as described in Section~\ref{sec:method:ann}, with different percentages of the dataset as stopping criterion.
The search results were then compared to the $k$ best results retrieved by linear search.
We chose different $k$ for comparison and took 200 samples and report statistics on recall.
This experiment should show that for a given fraction $\frac{c_{\max}}{n}$ of the dataset retrieving items in order of their Hamming distance results in significantly larger recall values than randomly selecting items which has an expected recall of $\frac{c_{\max}}{n}$.

Figure~\ref{fig:recall_by_hamming} presents the results with $\frac{c_{\max}}{n}$ in logarithmic steps on the x-axis and recall on the y-axis.
It can be seen that by iterating in order of the Hamming distance we only need to compare against a small fraction of the dataset to retrieve most of the nearest neighbors.
While there are error bars ranging from recall 0 to 1 for small percentages of the dataset increasing the search radius increases the recall to 1 and reduces errors significantly.
We also found the search parameter $k$ having little impact on the recall or search time, so $k$ was set to $100$ for the remaining experiments.

\subsection{Transverse Field}\label{sec:experiments:transverse}
\begin{figure}
    \centering
    \includegraphics[width=\linewidth]{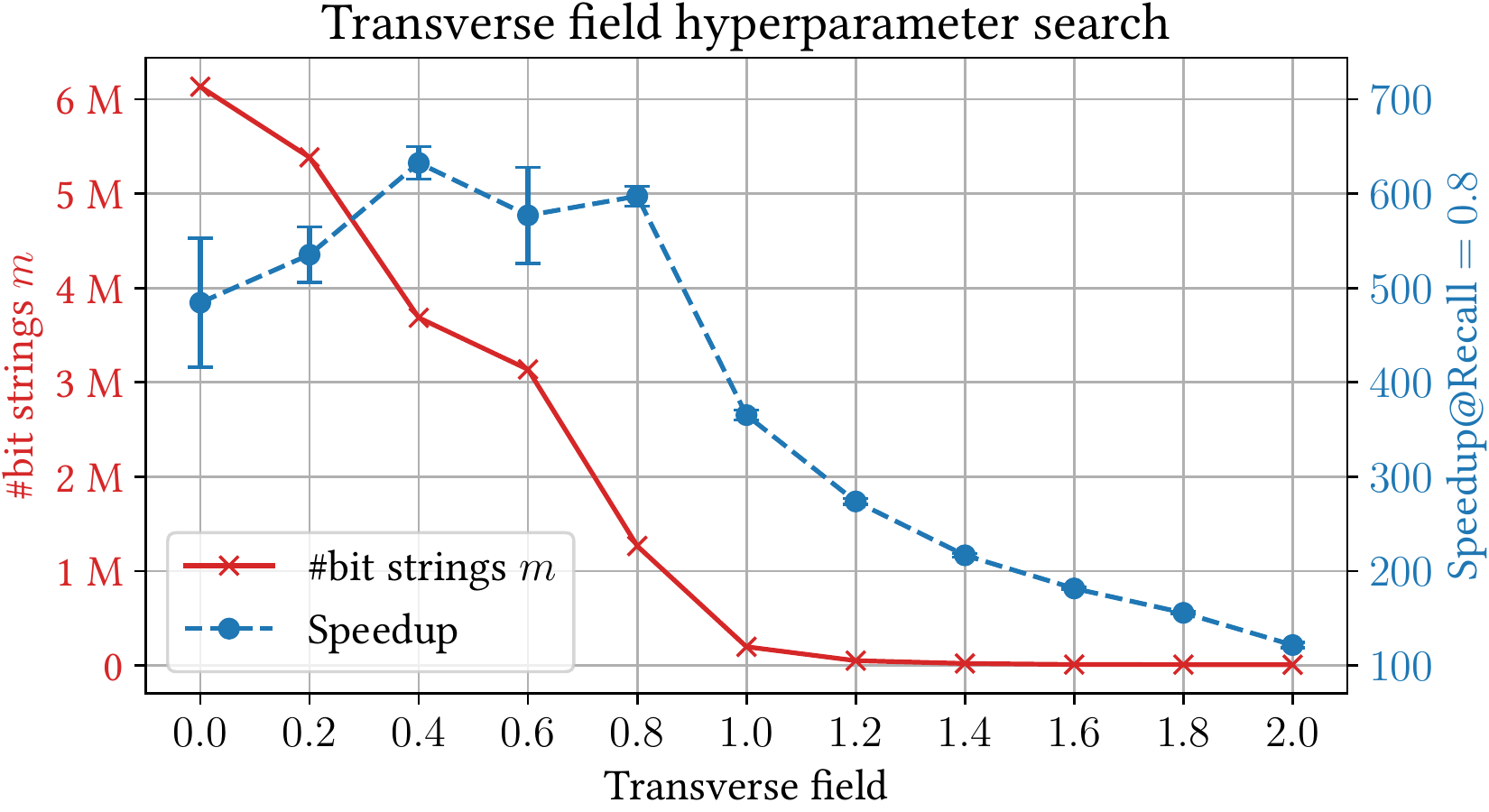}
    \caption{
        Impact of the transverse field on the latent space and search speed. 
        The red line shows the number of unique bit strings used to encode the dataset while the blue one shows the speedup at a fixed recall of $0.8$ for $k=100$. The error bars indicate the standard deviation on the speedup.
    }
    \label{fig:transverse_bins_speedup}
\end{figure}

As discussed in Section~\ref{sec:related:vae}, increasing the transverse field increases quantum fluctuations and narrows the learned distribution.
In the context of \gls{anns} this yields a hash function that maps to fewer distinct bit strings $m$.
In Section~\ref{sec:method:ann} we highlighted the trade-off between the query's speed and quality w.r.t. $m$.
In this experiment, we measured the effect of the transverse field in two ways 1) the number of bit strings $m$ and 2) the speedup at a fixed recall of 0.8.
We decided on using the simulated \gls{qbm} for this experiment for better reproducibility as there are many sources of noise in near-term quantum devices which we discuss further in section~\ref{sec:discussion}.
Additionally, at the time of writing this work, the performance of simulated sampling still outperformed the performance of an actual quantum sampler for small systems.

Both results can be found in Figure ~\ref{fig:transverse_bins_speedup} where the red line corresponds to the number of occupied bit strings and the blue line to the resulting speedup when performing \gls{anns}.
It can be seen that increasing the transverse field indeed narrows the learned distribution while the speedup shows a peak in the range of 0.4 to 0.8 supporting our assumption about the importance of the prior.
Instead of fixing a single value between these two, we opted for presenting results for both cases (0.4 and 0.8) in the next section to illustrate the behavior of the transverse field.

\subsection{Speedup}\label{sec:experiments:speedup}
\begin{figure}
    \centering
    \includegraphics[width=\linewidth]{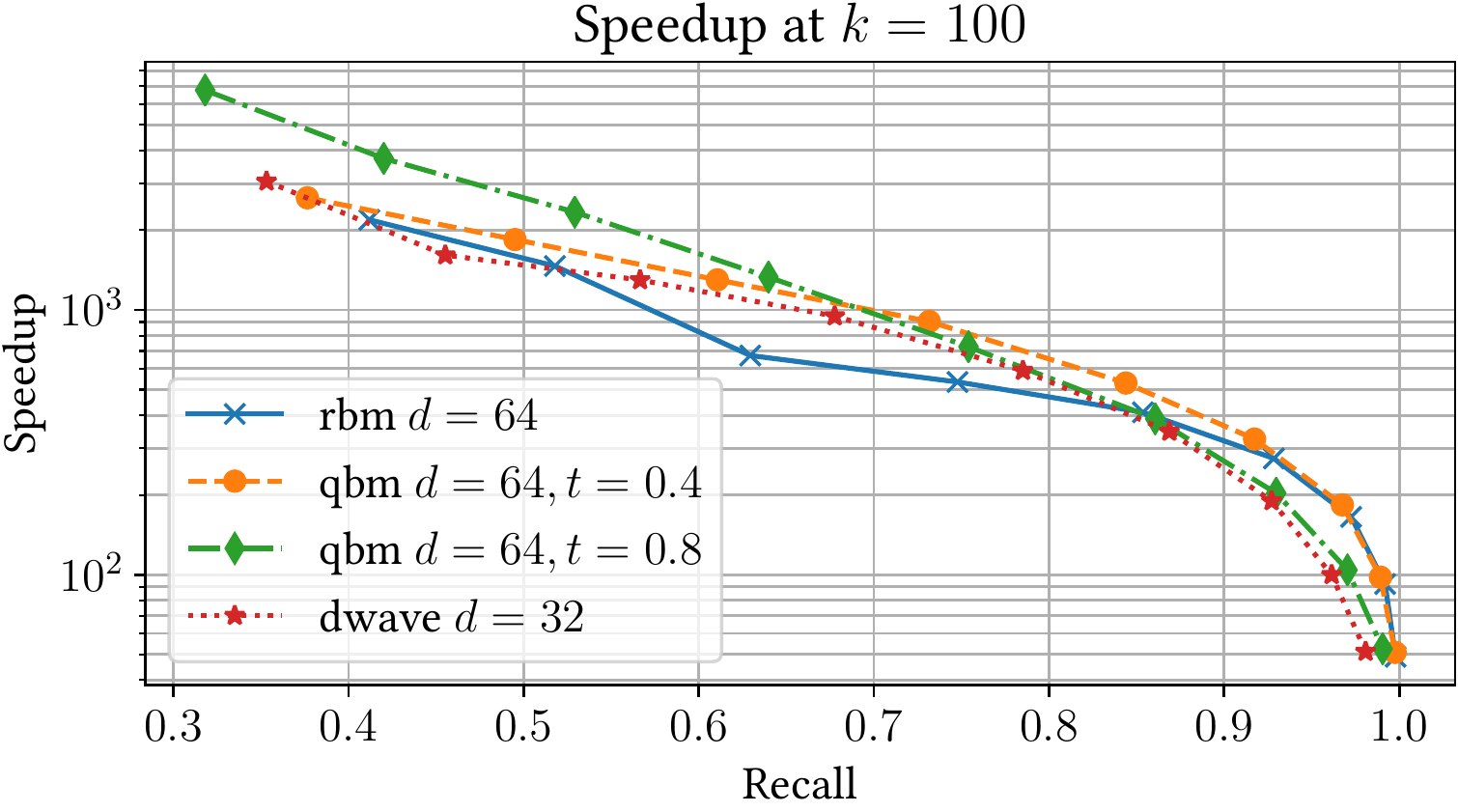}
    \caption{
        Speedup of models trained with a \gls{rbm}, simulated \gls{qbm} and annealed \gls{qbm} by recall for $k=100$. Closer to the top right corner is preferred.
    }
    \label{fig:speedup_by_recall}
\end{figure}

This experiment highlights the trade-off between query speed and quality and compares the different priors.
We choose 300 random data points from the dataset and computed the exact 100 nearest neighbors using linear search and compared them to the results from our \gls{anns} while measuring the wall time for both methods.

For the \gls{anns} we took fixed percentages of the dataset as stopping criterion, namely $\frac{c_{\max}}{n}\in\{0.01\%, 0.02\%, \hdots, 2.56\%\}$.
We did not include any larger nor smaller values as smaller values did not provide sufficient quality and higher values experience diminishing returns.
While we originally planned to compare all models with a 64bit latent space, we found differences between the simulated and hardware \gls{qbm}. 
The learned distribution widened when using the hardware model instead of narrowing as with the simulation.
Thus, we chose a model with a 32bit hardware \gls{qbm} such that the number of bit strings is approximately equal to the number of bit strings of the \gls{qbm} model with a transverse field of 0.4.
In addition to the hardware model, we present 64bit results of a \gls{rbm} as well as \glspl{qbm} with the transverse fields from Section~\ref{sec:experiments:transverse}, 0.4 and 0.8.

Figure~\ref{fig:speedup_by_recall} shows the trade-off between recall and speedup. 
For a fixed recall of 0.9, the annealed model performed worst with an average speedup of 263, while the classical provided a 326 times speedup and the simulated quantum models returned 373 and 283 times faster than linear search for transverse field 0.4 and 0.8, accordingly.
Relaxing the result quality to an average recall of 0.5 yields significant speedups for all models, for the quantum model we found an average speedup of 1485, the \gls{rbm} model returned 1587 times faster and the simulated \gls{qbm} models 1820 and 2714 times faster.
While the simulated quantum models yield the highest speedups for any given recall, the quantum annealed model performed strictly worse than both simulated models but resulted in higher speedups than the classical model for recall between 0.55 to 0.8.
We discuss more on the disparity between simulation and hardware in Section ~\ref{sec:discussion}. 

\subsection{Memory Consumption}\label{sec:experiments:memory}
\begin{table*}
    \centering
    \begin{tabular}{|c|c|c|c|c|c|}
        \hline
        Number of data points & \acrshort{qvae} & \acrshort{hnsw}@$M=6$ & \acrshort{hnsw}@$M=48$ & \acrshort{lsh}@$L=10$ & \acrshort{lsh}@$L=100$ \\
        \hline
        \hline
        4.465.537 (1 tile) & \bf 0.06 GB & 7.5 GB (0.6 GB) & 8.9 GB (2 GB) & 0.34 GB & 3.4 GB\\
        45.386.870 (USA) & \bf 0.59 GB & 73 GB (3 GB) & 88 GB (18 GB) & 3.4 GB & 34 GB\\
        113.789.007 (upper left quarter) & \bf 1.4 GB & 183 GB (8 GB) & 221 GB (46 GB) & 8.5 GB & 85 GB\\
        180.491.212 (left half) & \bf 2.6 GB & 291 GB (12 GB) & 350 GB (72 GB) & 14 GB & 135 GB\\
        534.748.234 (total) & \bf 6.5 GB & *860 GB (35 GB) & *1037 GB (212 GB) & 40 GB & *398 GB\\
        \hline
    \end{tabular}
    \caption{Memory usage of the search indices constructed by our method, \acrshort{hnsw} and \gls{lsh} for differently sized datasets. Numbers annotated by * are theoretical numbers where we were not able to actually construct the index due to memory limitations. Since \acrshort{hnsw} requires the complete dataset to be in memory, we calculated the space taken up by the actual index in brackets. }
    \label{tab:memory}
\end{table*}

In this section, we experimentally investigate the memory scaling discussed in Section~\ref{sec:method:ann}.
We compared our approach to the graph-based state-of-the-art
\gls{hnsw}\cite{malkov_efficient_2018}\footnote{\label{note:hnsw}We used the author's implementation on GitHub. \href{https://github.com/nmslib/hnswlib}{https://github.com/nmslib/hnswlib}} and \gls{lsh}.
\Gls{hnsw}'s memory usage is in $O(nM)$, \gls{lsh}'s memory usage is in $O(nL)$ and our approach is in $O(n)$ where $n$ is the number of elements, $M$ is the maximum node degree in \gls{hnsw} and $L$ is the number of hash tables in \gls{lsh}.
For a fixed $M$ and fixed $L$ all methods are in $O(n)$, though, our method has a significantly smaller pre-factor.
As discussed in Section~\ref{sec:method:ann}, our method requires only 8 bytes per object, equivalent to \gls{lsh} with a single hash table, while \gls{hnsw} requires $8M$ to $10M$ bytes per object \cite{malkov_efficient_2018} and \gls{lsh} requires $8L$ bytes per object.
Further, \gls{hnsw} needs to hold the complete dataset in memory to construct the index while our method and \gls{lsh} can process every data point independently to reduce memory usage during construction.
Here we show our simulated \gls{qbm} model with a transverse field of 0.4, \gls{hnsw} with the minimum and maximal recommended values for $M$, 6 and 48 respectively, and \gls{lsh} with commonly used lower and upper bounds, $L=10$ and $L=100$.
Though higher resolution hashes in \gls{lsh} require more space, we present optimistic estimates by choosing a low number of coarse hash functions per hash table, 10.
We evaluated different subsets of the \gls{modis} dataset to explore the limits of each method.

Table~\ref{tab:memory} shows the index sizes for each method for different dataset sizes.
Due to the requirement of \gls{hnsw} to access the complete dataset while constructing the index, we were not able to build the index for the complete \gls{modis} dataset as it exceeded the available memory.
Similarly, we were not able to build \gls{lsh} with $L=100$ for the complete \gls{modis} dataset as the index itself exceeded our available memory.
We found our method to scale favorably compared to \gls{hnsw} and \gls{lsh} as it enables the storage of the search index for the complete \gls{modis} dataset in memory and does not require the dataset to be loaded at build time.

\section{Discussion}\label{sec:discussion}
Although the results from Section~\ref{sec:experiments:latent} and Section~\ref{sec:experiments:transverse} back up our assumptions about the behavior of the latent space and the transverse field, Section~\ref{sec:experiments:speedup} shows different behavior when translating these insights to quantum hardware.
We found the hardware model to widen the learned distribution instead of narrowing it.
There are multiple factors that may cause this behavior.
Sources of noise include decoherence of the qubits as a result of coupling with the environment and thermal fluctuations. 
Moreover, there are errors from the mapping between logical variables and hardware qubits, in fact, an efficient mapping between embedded and logical space may not exist \cite{marshall_perils_2019}.
The latter error could be minimized by using \glspl{bm} which are native to the hardware, e.g., graphs with a maximum degree of 6.
Nonetheless, constantly improving hardware and algorithms will reduce noise and embedding errors over time.

Despite these shortcomings, the hardware and especially the simulated results show comparable performance to the \gls{rbm}, especially at lower recall values. Though this is a simple task, the data shown here is an encouraging sign that quantum hardware can be integrated into classical deep learning methods. 

As mentioned in Section~\ref{sec:experiments}, we excluded comparisons to other methods in Section~\ref{sec:experiments:speedup} as this work focuses on exploring similarity search as an early application for quantum computing by constructing a proof of concept.
There are a variety of unexplored options to improve upon our results, e.g., using more expressive encoders, including domain-specific features, adding similarity preserving losses \cite{do_learning_2016}, integrating faster Hamming searching and switching from Python to C++ like most \gls{anns} methods \cite{li_approximate_2016}.
However, here we provide some broader context to the field via a comparison to \gls{hnsw}\footnotemark[1].
At an average recall of 0.95, our best model resulted in an average speedup of 233 while \gls{hnsw} returned results around 200,603 times faster than linear search.
We used the best recommended parameter for $M$, 48, and determined $\mathit{ef}$ by a small hyperparameter search, as recommended \cite{malkov_efficient_2018}, and fixed it to 400 for construction and 100 during search.

The last point we want to address here is the memory issue of \gls{hnsw} and \gls{lsh} from Section~\ref{sec:experiments:memory}.
It should be mentioned that workarounds to the memory issues for \gls{hnsw} and \gls{lsh} exists, e.g., offloading the index to non-volatile memory but this would significantly slow down both methods.
In contrast to that, our approach is capable of keeping an index for half a billion data points in main memory.

\section{Conclusion}\label{sec:conclusion}
To summarize, we explored a \gls{qvae}-based approach to similarity search in large-scale high-dimensional datasets as an early application target for quantum computing.
We discussed several potential advantages of our approach:
The data is inherently binarized, the memory scales favorably, quantum fluctuations in form of the transverse field can be used to gain continuous control over the shape of the latent space and the training of the \gls{qvae} can be assisted using quantum annealers.
We experimentally demonstrated the approximation quality of the Hamming distance in the latent space to the Euclidean distance in the original space and verified real-world speedups on the \gls{modis} dataset with low space requirements.

These insights may gain importance with increasing complexity of quantum devices, as they become intractable to simulate \cite{arute2019quantum}.
Whether this means that quantum devices will exceed classical in application is an open question, one we began addressing here.

\section*{Acknowledgments}
We would like to thank Scott D. Pakin from Los Alamos National Laboratory (LANL) for his help with running the experiments and are grateful for support from NASA's Advanced Information Systems Technology Program (grant \#AIST16-0137) and NASA Ames Research Center. We appreciate support from the AFRL Information Directorate under grant F4HBKC4162G001 and the Office of the Director of National Intelligence (ODNI) and the Intelligence Advanced Research Projects Activity (IARPA), via IAA 145483.  The views and conclusions contained herein are those of the authors and should not be interpreted as necessarily representing the official policies or endorsements, either expressed or implied, of ODNI, IARPA, AFRL, or the U.S. Government. 
The U.S. Government is authorized to reproduce and distribute reprints for Governmental purpose notwithstanding any copyright annotation thereon.

\bibliography{bibliography.bib} 

\begin{thebibliography}{10}

\bibitem{biamonte2017quantum}
J.~Biamonte, P.~Wittek, N.~Pancotti, P.~Rebentrost, N.~Wiebe, and S.~Lloyd,
  ``Quantum machine learning,'' {\em Nature}, vol.~549, no.~7671, pp.~195--202,
  2017.

\bibitem{lloyd2014quantum}
S.~Lloyd, M.~Mohseni, and P.~Rebentrost, ``Quantum principal component
  analysis,'' {\em Nature Physics}, vol.~10, no.~9, pp.~631--633, 2014.

\bibitem{rebentrost2014quantum}
P.~Rebentrost, M.~Mohseni, and S.~Lloyd, ``Quantum support vector machine for
  big data classification,'' {\em Physical review letters}, vol.~113, no.~13,
  p.~130503, 2014.

\bibitem{mott2017solving}
A.~Mott, J.~Job, J.-R. Vlimant, D.~Lidar, and M.~Spiropulu, ``Solving a higgs
  optimization problem with quantum annealing for machine learning,'' {\em
  Nature}, vol.~550, no.~7676, pp.~375--379, 2017.

\bibitem{vinci2019path}
W.~Vinci, L.~Buffoni, H.~Sadeghi, A.~Khoshaman, E.~Andriyash, and M.~H. Amin,
  ``A path towards quantum advantage in training deep generative models with
  quantum annealers,'' {\em arXiv preprint arXiv:1912.02119}, 2019.

\bibitem{khoshaman2018quantum}
A.~Khoshaman, W.~Vinci, B.~Denis, E.~Andriyash, and M.~H. Amin, ``Quantum
  variational autoencoder,'' {\em Quantum Science and Technology}, vol.~4,
  no.~1, p.~014001, 2018.

\bibitem{rolfe_discrete_2016}
J.~T. Rolfe, ``Discrete {Variational} {Autoencoders},'' {\em arXiv:1609.02200
  [cs, stat]}, Sept. 2016.
\newblock arXiv: 1609.02200.

\bibitem{preskill2018quantum}
J.~Preskill, ``Quantum computing in the nisq era and beyond,'' {\em Quantum},
  vol.~2, p.~79, 2018.

\bibitem{arute2019quantum}
F.~Arute, K.~Arya, R.~Babbush, D.~Bacon, J.~C. Bardin, R.~Barends, R.~Biswas,
  S.~Boixo, F.~G. Brandao, D.~A. Buell, {\em et~al.}, ``Quantum supremacy using
  a programmable superconducting processor,'' {\em Nature}, vol.~574, no.~7779,
  pp.~505--510, 2019.

\bibitem{kingma_auto-encoding_2013}
D.~P. Kingma and M.~Welling, ``Auto-encoding variational bayes,'' {\em arXiv
  preprint arXiv:1312.6114}, 2013.

\bibitem{amin2018quantum}
M.~H. Amin, E.~Andriyash, J.~Rolfe, B.~Kulchytskyy, and R.~Melko, ``Quantum
  boltzmann machine,'' {\em Physical Review X}, vol.~8, no.~2, p.~021050, 2018.

\bibitem{long2010restricted}
P.~Long and R.~Servedio, ``Restricted {{Boltzmann Machines}} are {{Hard}} to
  {{Approximately Evaluate}} or {{Simulate}},'' in {\em {{ICML}} 2010 -
  {{Proceedings}}, 27th {{International Conference}} on {{Machine Learning}}},
  pp.~703--710, Aug. 2010.

\bibitem{li_approximate_2016}
W.~Li, Y.~Zhang, Y.~Sun, W.~Wang, W.~Zhang, and X.~Lin, ``Approximate {{Nearest
  Neighbor Search}} on {{High Dimensional Data}} --- {{Experiments}},
  {{Analyses}}, and {{Improvement}} (v1.0),'' {\em arXiv:1610.02455 [cs]}, Oct.
  2016.

\bibitem{camerra2014beyond}
A.~Camerra, J.~Shieh, T.~Palpanas, T.~Rakthanmanon, and E.~Keogh, ``Beyond one
  billion time series: indexing and mining very large time series collections
  with $i$sax2+,'' {\em Knowledge and information systems}, vol.~39, no.~1,
  pp.~123--151, 2014.

\bibitem{johnson_billion-scale_2017}
J.~Johnson, M.~Douze, and H.~J{\'e}gou, ``Billion-scale similarity search with
  {{GPUs}},'' {\em arXiv:1702.08734 [cs]}, Feb. 2017.

\bibitem{yagoubi_radiussketch:_2017}
D.~E. Yagoubi, R.~Akbarinia, F.~Masseglia, and D.~Shasha, ``{RadiusSketch}:
  Massively distributed indexing of time series,'' in {\em 2017 {IEEE}
  International Conference on Data Science and Advanced Analytics ({DSAA})},
  pp.~262--271, {IEEE}, 2017.

\bibitem{do_learning_2016}
T.-T. Do, A.-D. Doan, and N.-M. Cheung, ``Learning to {Hash} with {Binary}
  {Deep} {Neural} {Network},'' {\em arXiv:1607.05140 [cs]}, July 2016.
\newblock arXiv: 1607.05140.

\bibitem{srivastava_unsupervised_2015}
N.~Srivastava, E.~Mansimov, and R.~Salakhutdinov, ``Unsupervised {Learning} of
  {Video} {Representations} using {LSTMs},'' {\em arXiv:1502.04681 [cs]}, Feb.
  2015.
\newblock arXiv: 1502.04681.

\bibitem{carreira-perpinan_hashing_2015}
M.~A. Carreira-Perpinan and R.~Raziperchikolaei, ``Hashing with binary
  autoencoders,'' in {\em 2015 {IEEE} {Conference} on {Computer} {Vision} and
  {Pattern} {Recognition} ({CVPR})}, (Boston, MA, USA), pp.~557--566, IEEE,
  June 2015.

\bibitem{cao_correlation_2016}
Y.~Cao, M.~Long, J.~Wang, and H.~Zhu, ``Correlation autoencoder hashing for
  supervised cross-modal search,'' in {\em Proceedings of the 2016 {ACM} on
  International Conference on Multimedia Retrieval - {ICMR} '16}, pp.~197--204,
  {ACM} Press, 2016.

\bibitem{board2007earth}
S.~S. Board, N.~R. Council, {\em et~al.}, {\em Earth science and applications
  from space: national imperatives for the next decade and beyond}.
\newblock National Academies Press, 2007.

\bibitem{wright2002automated}
R.~Wright, L.~Flynn, H.~Garbeil, A.~Harris, and E.~Pilger, ``Automated volcanic
  eruption detection using modis,'' {\em Remote sensing of environment},
  vol.~82, no.~1, pp.~135--155, 2002.

\bibitem{zhang2003monitoring}
X.~Zhang, M.~A. Friedl, C.~B. Schaaf, A.~H. Strahler, J.~C. Hodges, F.~Gao,
  B.~C. Reed, and A.~Huete, ``Monitoring vegetation phenology using modis,''
  {\em Remote sensing of environment}, vol.~84, no.~3, pp.~471--475, 2003.

\bibitem{remer2005modis}
L.~A. Remer, Y.~Kaufman, D.~Tanr{\'e}, S.~Mattoo, D.~Chu, J.~V. Martins, R.-R.
  Li, C.~Ichoku, R.~Levy, R.~Kleidman, {\em et~al.}, ``The modis aerosol
  algorithm, products, and validation,'' {\em Journal of the atmospheric
  sciences}, vol.~62, no.~4, pp.~947--973, 2005.

\bibitem{boriah2008land}
S.~Boriah, V.~Kumar, C.~Potter, M.~Steinbach, and S.~Klooster, ``Land cover
  change detection using data mining techniques,'' {\em Technical Report March
  14}, 2008.

\bibitem{steinhaeuser2012multivariate}
K.~Steinhaeuser, A.~R. Ganguly, and N.~V. Chawla, ``Multivariate and multiscale
  dependence in the global climate system revealed through complex networks,''
  {\em Climate dynamics}, vol.~39, no.~3-4, pp.~889--895, 2012.

\bibitem{kamath2001mining}
C.~Kamath, ``On mining scientific datasets,'' in {\em Data Mining for
  Scientific and Engineering Applications}, pp.~1--21, Springer, 2001.

\bibitem{zhou2015teleconnection}
D.~Zhou, A.~Gozolchiani, Y.~Ashkenazy, and S.~Havlin, ``Teleconnection paths
  via climate network direct link detection,'' {\em Physical review letters},
  vol.~115, no.~26, p.~268501, 2015.

\bibitem{indyk_approximate_1998}
P.~Indyk and R.~Motwani, ``Approximate nearest neighbors: towards removing the
  curse of dimensionality,'' in {\em Proceedings of the thirtieth annual {ACM}
  symposium on Theory of computing}, pp.~604--613, {ACM}, 1998.

\bibitem{gionis_similarity_1999}
A.~Gionis, P.~Indyk, and R.~Motwani, ``Similarity search in high dimensions via
  hashing,'' in {\em Vldb}, vol.~99, pp.~518--529, 1999.

\bibitem{hoffman_stochastic_2013}
M.~D. Hoffman, D.~M. Blei, C.~Wang, and J.~Paisley, ``Stochastic variational
  inference,'' {\em The Journal of Machine Learning Research}, vol.~14, no.~1,
  pp.~1303--1347, 2013.

\bibitem{khoshaman2018gumbolt}
A.~H. Khoshaman and M.~Amin, ``Gumbolt: extending gumbel trick to boltzmann
  priors,'' in {\em Advances in Neural Information Processing Systems},
  pp.~4061--4070, 2018.

\bibitem{hoffman_elbo_2016}
M.~D. Hoffman and M.~J. Johnson, ``Elbo surgery: Yet another way to carve up
  the variational evidence lower bound,'' in {\em Workshop in {{Advances}} in
  {{Approximate Bayesian Inference}}, {{NIPS}}}, vol.~1, p.~2, 2016.

\bibitem{so_nderby_ladder_2016}
C.~K. {S{\o} nderby}, T.~Raiko, L.~{Maal{\o} e}, S.~r.~K. {S{\o} nderby}, and
  O.~Winther, ``Ladder {{Variational Autoencoders}},'' in {\em Advances in
  {{Neural Information Processing Systems}} 29} (D.~D. Lee, M.~Sugiyama, U.~V.
  Luxburg, I.~Guyon, and R.~Garnett, eds.), pp.~3738--3746, {Curran Associates,
  Inc.}, 2016.

\bibitem{casale_gaussian_2018}
F.~P. Casale, A.~Dalca, L.~Saglietti, J.~Listgarten, and N.~Fusi, ``Gaussian
  {{Process Prior Variational Autoencoders}},'' in {\em Advances in {{Neural
  Information Processing Systems}} 31} (S.~Bengio, H.~Wallach, H.~Larochelle,
  K.~Grauman, N.~{Cesa-Bianchi}, and R.~Garnett, eds.), pp.~10369--10380,
  {Curran Associates, Inc.}, 2018.

\bibitem{hinton_training_2002}
G.~E. Hinton, ``Training products of experts by minimizing contrastive
  divergence,'' {\em Neural computation}, vol.~14, no.~8, pp.~1771--1800, 2002.

\bibitem{tieleman2009using}
T.~Tieleman and G.~Hinton, ``Using fast weights to improve persistent
  contrastive divergence,'' in {\em Proceedings of the 26th Annual
  International Conference on Machine Learning}, pp.~1033--1040, 2009.

\bibitem{koller2007graphical}
D.~Koller, N.~Friedman, L.~Getoor, and B.~Taskar, ``Graphical models in a
  nutshell,'' {\em Introduction to statistical relational learning},
  pp.~13--55, 2007.

\bibitem{rieffel2019ans}
E.~G. Rieffel, S.~Hadfield, T.~Hogg, S.~Mandr{\`a}, J.~Marshall, G.~Mossi,
  B.~O'Gorman, E.~Plamadeala, N.~M. Tubman, D.~Venturelli, {\em et~al.}, ``From
  ans{\"a}tze to z-gates: a {NASA} view of quantum computing,'' {\em arXiv
  preprint arXiv:1905.02860}, 2019.

\bibitem{ronnow2014defining}
T.~F. R{\o}nnow, Z.~Wang, J.~Job, S.~Boixo, S.~V. Isakov, D.~Wecker, J.~M.
  Martinis, D.~A. Lidar, and M.~Troyer, ``Defining and detecting quantum
  speedup,'' {\em Science}, vol.~345, no.~6195, pp.~420--424, 2014.

\bibitem{katzgraber2014glassy}
H.~G. Katzgraber, F.~Hamze, and R.~S. Andrist, ``Glassy chimeras could be blind
  to quantum speedup: Designing better benchmarks for quantum annealing
  machines,'' {\em Physical Review X}, vol.~4, no.~2, p.~021008, 2014.

\bibitem{biswas2017nasa}
R.~Biswas, Z.~Jiang, K.~Kechezhi, S.~Knysh, S.~Mandra, B.~O’Gorman,
  A.~Perdomo-Ortiz, A.~Petukhov, J.~Realpe-G{\'o}mez, E.~Rieffel, {\em et~al.},
  ``A {NASA} perspective on quantum computing: Opportunities and challenges,''
  {\em Parallel Computing}, vol.~64, pp.~81--98, 2017.

\bibitem{adachi2015application}
S.~H. Adachi and M.~P. Henderson, ``Application of quantum annealing to
  training of deep neural networks,'' {\em arXiv preprint arXiv:1510.06356},
  2015.

\bibitem{kingma_adam_2017}
D.~P. Kingma and J.~Ba, ``Adam: {{A Method}} for {{Stochastic Optimization}},''
  {\em arXiv:1412.6980 [cs]}, Jan. 2017.

\bibitem{gull2011continuous}
E.~Gull, A.~J. Millis, A.~I. Lichtenstein, A.~N. Rubtsov, M.~Troyer, and
  P.~Werner, ``Continuous-time monte carlo methods for quantum impurity
  models,'' {\em Reviews of Modern Physics}, vol.~83, no.~2, p.~349, 2011.

\bibitem{justice1998moderate}
C.~O. Justice, E.~Vermote, J.~R. Townshend, R.~Defries, D.~P. Roy, D.~K. Hall,
  V.~V. Salomonson, J.~L. Privette, G.~Riggs, A.~Strahler, {\em et~al.}, ``The
  moderate resolution imaging spectroradiometer (modis): Land remote sensing
  for global change research,'' {\em IEEE transactions on geoscience and remote
  sensing}, vol.~36, no.~4, pp.~1228--1249, 1998.

\bibitem{solano2010modis}
R.~Solano, K.~Didan, A.~Jacobson, and A.~Huete, ``Modis vegetation index
  user’s guide (mod13 series),'' {\em Vegetation Index and Phenology Lab, The
  University of Arizona}, pp.~1--38, 2010.

\bibitem{malkov_efficient_2018}
Y.~A. Malkov and D.~A. Yashunin, ``Efficient and robust approximate nearest
  neighbor search using {{Hierarchical Navigable Small World}} graphs,'' {\em
  arXiv:1603.09320 [cs]}, Aug. 2018.

\bibitem{marshall_perils_2019}
J.~Marshall, A.~Di~Gioacchino, and E.~G. Rieffel, ``The {{Perils}} of
  {{Embedding}} for {{Sampling Problems}},'' {\em arXiv:1909.12184 [cond-mat,
  physics:quant-ph]}, Sept. 2019.

\end{thebibliography}
\bibliographystyle{ieeetr}

\end{document}